\documentclass[11pt]{article}

\usepackage[utf8]{inputenc}
\usepackage[T1]{fontenc}
\usepackage[a4paper,margin=2.35cm]{geometry}
\usepackage{amsmath,amssymb}
\usepackage{booktabs}
\usepackage{array}
\usepackage{graphicx}
\usepackage{float}
\usepackage{xcolor}
\usepackage{xspace}
\usepackage[round,authoryear]{natbib}
\usepackage[colorlinks=true,allcolors=blue!55!black]{hyperref}

\graphicspath{
  {figures/}
}

\newcommand{\LSMone}{LSM-1\xspace}
\newcommand{\LSMtwo}{LSM-2\xspace}
\newcommand{\ASM}{ASM\xspace}

\title{Distribution-free false-alarm calibration and chance-corrected spatial evaluation for industrial anomaly detection}
\author{Jie Deng\thanks{Corresponding author: \href{mailto:dengjie.work@foxmail.com}{dengjie.work@foxmail.com}; ORCID: \href{https://orcid.org/0009-0001-3470-2049}{0009-0001-3470-2049}.}\\
\small Tongji University}
\date{}

\begin{document}
\maketitle

\begin{abstract}
Studies of industrial visual inspection commonly report the area under the receiver operating characteristic curve (AUROC) and the overlap between anomaly maps and defect masks. Neither measure specifies the false-alarm rate at a selected threshold, while recurrent defect locations and mask geometry can inflate overlap. We combine a distribution-free upper tolerance threshold with a paired-minus-crossed spatial test. This test compares each detector's score-contributing locations with the matched defect mask and with masks from other images; the difference in rates defines spatial-evidence lift relative to the empirical chance-overlap rate. We evaluate three detectors on 120 point-defect images from three ISP-AD modalities and three fixed data splits. Of 378 alarms, 230 overlap the matched mask. Paired and crossed rates are nevertheless similar in eight of nine detector--modality cells; only DINOv2--ASM has a positive 95\% bootstrap lower bound (lift 0.259, 95\% interval 0.159--0.347). On the independent Magnetic Tile Defect dataset, the same analysis gives lifts of 0.203 (0.169--0.236) for Wide ResNet-50 (WRN50) patch memory and 0.231 (0.202--0.262) for Vision Transformer B/16 (ViT-B/16) patch memory, with one-sided permutation $p=10^{-5}$ for both. When crossed masks are restricted to the same defect class, the lifts remain 0.185 and 0.210. Exact sample planning shows that, with 150 calibration normals, a 95\%-confidence distribution-free claim is supported only for target false-positive rates of 1.98\% or higher; a 1\% target requires at least 299 normals. The results support reporting operating-point performance and chance-corrected spatial evidence alongside AUROC and raw mask overlap.
\end{abstract}


\section{Introduction}
\label{sec:introduction}

Automated visual inspection supports manufacturing quality control when manual inspection is costly, slow, or difficult to standardize. In unsupervised applications, anomaly detectors are commonly evaluated as ranking methods. A detector assigns an image or pixel score, and AUROC summarizes its ranking performance on datasets such as MVTec AD and VisA~\citep{bergmann2021mvtec,zou2022visa}. An inspection system also needs an operating threshold, an estimate of the associated false-positive rate (FPR), and the defect recall at that threshold. These quantities cannot be recovered from AUROC alone. A detector may rank defects above normal images while still missing many defects at the selected operating point.

Localization introduces a second evaluation problem, particularly for small defects. A common check asks whether the maximum or top-$k$ scoring location intersects the annotated defect mask. The overlap is easy to calculate, but it is affected by centered crops, recurrent defect positions, mask dilation, the number of selected patches, and mask area. A detector can therefore obtain a high overlap rate by responding to a common image region rather than to an image-specific defect. A useful spatial evaluation should compare the observed image--mask pair with a reference that retains the score maps and mask geometry while breaking the original correspondence.

We evaluate anomaly detectors at an explicit operating point. A one-sided order-statistic tolerance limit is fitted to target-domain normal scores, and a paired-minus-crossed permutation test estimates spatial evidence beyond the dataset's empirical chance-overlap level. The primary experiments use three optical modalities from ISP-AD~\citep{krassnig2026ispad}, three detector families, and three support/calibration splits. We then apply a protocol fixed before result inspection to the independent Magnetic Tile Defect dataset~\citep{huang2020magnetictile}, using convolutional and transformer patch memories.

The study makes three contributions.

\begin{enumerate}
  \item We link anomaly scores to a finite-sample, distribution-free false-positive statement and derive the normal-sample requirements for target FPRs between 0.1\% and 10\%.
  \item We distinguish image alarms, raw mask overlap, and chance-corrected spatial-evidence lift. The crossed-mask reference distribution preserves the observed mask geometry and detector locations while removing the original image--mask pairing.
  \item We evaluate the procedure on ISP-AD and Magnetic Tile Defect. The comparison includes convolutional and transformer representations, repeated data splits, and class- and filename-group-matched sensitivity analyses.
\end{enumerate}

The detector architectures are not modified. The aim is to make threshold selection and spatial interpretation auditable while keeping ranking performance, operating-point performance, and mask overlap distinct. Appendix~\ref{app:efficientad} reports a secondary EfficientAD reliability check and shows how a run with non-finite scores is retained in the record.

\section{Related work}
\label{sec:related}

\subsection{Industrial anomaly detectors and datasets}

PatchCore scores pretrained convolutional patches by distance to a nominal memory~\citep{roth2022patchcore}; PaDiM models a position-wise Gaussian distribution~\citep{defard2021padim}. DINOv2 provides transferable self-supervised transformer features~\citep{oquab2024dinov2}, and AnomalyDINO demonstrates their utility for few-shot anomaly detection~\citep{damm2025anomalydino}. EfficientAD combines a compact teacher--student branch with an autoencoder branch for low-latency inspection~\citep{batzner2024efficientad}; Anomalib supplies a maintained implementation framework~\citep{akcay2022anomalib}. Rather than modifying these detectors, we examine the evidence provided by their outputs at a stated operating point.

MVTec AD and VisA established broad image- and pixel-level benchmarks~\citep{bergmann2021mvtec,zou2022visa}. ISP-AD contributes real production defects observed through three optical modalities~\citep{krassnig2026ispad}; Magnetic Tile Defect supplies five defect types with pixel masks and a large defect-free class~\citep{huang2020magnetictile}. MVTec AD 2 emphasizes the difficulty of maintaining thresholds under industrial distribution change~\citep{hecklerkram2026mvtecad2}.

\subsection{Operating thresholds and detection reliability}

Threshold-free ranking cannot specify the alarm rate of a deployed system~\citep{cools2026threshold}. A one-sided order-statistic tolerance limit bounds the upper tail without a parametric model for normal scores~\citep{wilks1941tolerance}. Probability-of-detection studies similarly treat detection as a binary event at a chosen threshold and report uncertainty conditional on flaw properties~\citep{astm2023e2862,falcetelli2022reliability}. The threshold used here has a tolerance-limit interpretation and differs from the marginal guarantee of split conformal prediction.

\subsection{Spatial evaluation and empirical chance}

Pixel AUROC, region overlap, and point-mask intersection quantify different spatial properties. For small defects, intersection depends not only on a detector's response to the defect but also on how many locations are selected and where masks tend to occur. Metric choice and spatial bias are established concerns in visual-saliency evaluation~\citep{bylinskii2019saliency}. In particular, shuffled AUC samples negative locations from other images to reduce center bias, although this construction can still retain central or peripheral biases~\citep{jia2020fnaauc}.

Cross-image spatial negatives provide a useful reference in saliency evaluation. We adapt this idea to a thresholded, unconditional alarm-and-mask event for industrial anomaly detection. The analysis uses complete defect masks and detector-native contributor sets rather than fixation negatives or a threshold-free AUC. Pairing a score map with masks from other defect images retains the detector's spatial pattern, alarm frequency, and empirical mask geometry. Image-level resampling and assignment permutations quantify uncertainty at the selected operating point.

\section{Materials and methods}
\label{sec:methods}

\subsection{Analysis status and temporal ordering}

The ISP-AD operating-point and crossed-mask analyses form the primary study. After completing that analysis, we fixed the Magnetic Tile protocol before inspecting its results. Magnetic Tile is therefore an independent-dataset evaluation of the revised procedure, but not a preregistration of the complete study. Pooled crossing is the primary external statistic; the class-stratified and class--filename-group-stratified variants are post-hoc sensitivity analyses. The defect-covariate models were planned earlier, although their use as mask-geometry diagnostics followed the crossed-mask analysis. The EfficientAD experiment in Appendix~\ref{app:efficientad} was conducted later as a secondary score-validity check.

\subsection{Datasets, data roles, and analysis units}

The primary evaluation uses the public unsupervised split of ISP-AD. The native labels \LSMtwo, \LSMone, and \ASM denote distinct optical acquisition modalities. Point defects form the spatial-analysis population because their small extent makes accidental overlap and patch-grid resolution particularly important. AUROC includes all native anomaly types.

\begin{table}[H]
\centering
\caption{Normal-data roles and defect counts. Roles are disjoint within seed. ISP counts are point defects/all anomalies. MTD image metrics include all 392 defects; four empty masks are excluded only from spatial inference.}
\label{tab:data}
\footnotesize
\begin{tabular}{@{}llrrrrr@{}}
\toprule
Dataset & Modality / model & Support & Calib. & Eval. normal & Defects & Spatial $n$ \\
\midrule
ISP-AD & \LSMtwo & 300 & 150 & 200 & 33 / 63 & 33 \\
ISP-AD & \LSMone & 300 & 150 & 200 & 46 / 49 & 46 \\
ISP-AD & \ASM    & 300 & 150 & 200 & 41 / 33 & 41 \\
\addlinespace
Magnetic Tile & WRN50 / ViT-B16 & 16 & 150 & 786 & 392 / 392 & 388 \\
\bottomrule
\end{tabular}
\end{table}

Three deterministic seeds (20260801--20260803) define support, calibration, and independent-normal roles. The 120 ISP-AD point-defect files are modality-specific images; the dataset documentation does not establish a cross-modality physical-specimen identifier, so we do not merge same-numbered files across modalities. Each image is evaluated under three detectors and three seeds, yielding $120\times3\times3=1080$ repeated evaluations. Inference and resampling use the full image path as the cluster key.

We specified the Magnetic Tile protocol before examining its spatial results. Per seed, it uses 16 support normals, 150 calibration normals, 786 independent normals, and every defect image. Its five defect classes contain 392 images; four masks with no positive pixel are reported and excluded from spatial calculations. No defect image is used to fit a detector or threshold in either dataset.

\subsection{Detector configurations}
\label{sec:detectors}

The ISP-AD comparison uses three fixed feature-based detectors. Exact-memory PatchCore retains all Wide-ResNet-50 layer-2/layer-3 support patches at $256\times256$ input and scores an image by the maximum $k=1$ nearest-memory distance. DINOv2 patch memory uses a ViT-S/14 with fixed pretrained weights at $252\times252$, cosine distance, and the mean of the largest 1\% patch scores. PaDiM-style modeling aligns the first three Wide-ResNet-50 stages, retains 100 seed-fixed channels, applies ridge $0.01$, and averages the largest 1\% Mahalanobis distances.

The Magnetic Tile evaluation uses exact patch memories with either Wide-ResNet-50 layer-2/layer-3 features or ViT-B/16 final tokens at $224\times224$. All patches from 16 support normals are retained. For both detectors, the image score and spatial contributor are determined by the maximum nearest-memory distance. Masks, anomaly labels, and evaluation results are not used to configure the detectors.

\subsection{Distribution-free operating threshold}
\label{sec:calibration}

Let $S_1,\ldots,S_m$ be independent and identically distributed scores from a fixed target-normal distribution $P_0$, after all elements of the scoring pipeline have been specified. Write $S_{(1)}\leq\cdots\leq S_{(m)}$. For target upper-tail probability $\alpha$ and calibration confidence $1-\delta$, choose the least conservative rank $r$ satisfying
\begin{equation}
  \Pr\{Z\geq m-r+1\}\geq1-\delta,
  \qquad Z\sim\mathrm{Binomial}(m,\alpha),
  \label{eq:binomial}
\end{equation}
and set $\tau=S_{(r)}$. The deployed decision is the strict rule $S>\tau$. The following guarantee also holds for discrete score distributions and ties:
\begin{equation}
  \Pr_{\mathrm{cal}}
  \left[P_0\{S>\tau\}\leq\alpha\right]\geq1-\delta.
  \label{eq:guarantee}
\end{equation}
Appendix~\ref{app:proof} gives a short proof. The binomial calculation requires independent calibration units. Its coverage is conditional on a fixed normal distribution and does not extend to batch dependence or domain shift.

The primary setting is $m=150$, $\alpha=0.10$, and $\delta=0.05$, which gives $r=142$ and achieved confidence 0.9693. Two hundred independent ISP-AD normals per modality and seed, and 786 Magnetic Tile normals per seed, estimate realized FPR with Wilson intervals~\citep{wilson1927interval}; they do not refit the threshold.

\subsection{From raw overlap to chance-corrected spatial evidence}
\label{sec:spatial}

For defect image $j$, detector $d$, modality $m$, and seed $s$, define the image alarm as
\begin{equation}
  A_{jdms}=\mathbb{1}\{S_{jdms}>\tau_{dms}\}
\end{equation}
Let $G_{jdms}$ denote the score-contributing grid cells: the maximum cell for PatchCore, the top 1\% cells for DINOv2 and PaDiM, and the maximum cell for both Magnetic Tile detectors. Each defect mask is mapped to the detector grid by any-positive pooling and dilated by one feature cell. For mask $i$ and score map $j$, define
\begin{equation}
  C^{(s)}_{ij}=A_{jdms}\,
  \mathbb{1}\{G_{jdms}\cap M_i^{+}\neq\varnothing\}.
  \label{eq:compatibility}
\end{equation}
The diagonal $C_{ii}^{(s)}$ is a \emph{raw alarm-and-overlap event}. We use this descriptive term because intersection with the corresponding mask has not yet been adjusted for chance overlap.

Within each detector--modality cell, the seed matrices are averaged:
\begin{equation}
  \bar C=3^{-1}\sum_s C^{(s)}.
\end{equation}
For $n$ defect images, the paired rate, crossed rate, and spatial-evidence lift are
\begin{align}
  \hat p_{\mathrm{pair}} &= \frac{1}{n}\operatorname{tr}(\bar C), \\
  \hat p_{\mathrm{cross}} &=
  \frac{\sum_{i,j}\bar C_{ij}-\operatorname{tr}(\bar C)}{n(n-1)}, \\
  \hat\Delta &= \hat p_{\mathrm{pair}}-\hat p_{\mathrm{cross}}.
  \label{eq:lift}
\end{align}
Crossing preserves the observed score maps, alarm decisions, mask sizes, and mask locations but breaks image--mask identity. A 5,000-replicate paired-image bootstrap gives a percentile interval for $\Delta$. A one-sided 99,999-permutation test randomly assigns masks to maps. ISP-AD cellwise tests are interpreted with Bonferroni threshold $0.05/9$; detector-level blocked permutations shuffle separately within modality. The prespecified external criterion requires positive bootstrap lower bounds and $p<0.025$ for both detector families.

The primary external statistic, specified in advance, crosses masks across all 388 nonempty-mask images. Because defect classes can differ in both geometry and typical position, we also perform two post-hoc sensitivity analyses. If $g(i)$ denotes a stratum, the crossed term is replaced by the image-weighted within-stratum rate
\begin{equation}
 \hat p_{\mathrm{cross},g}=\frac{1}{n}\sum_h
 \frac{\sum_{i\neq j:g(i)=g(j)=h}\bar C_{ij}}{n_h-1}.
 \label{eq:stratified-crossed}
\end{equation}
Masks are permuted and images bootstrapped only within (i) defect class and (ii) defect class crossed with the filename-defined \texttt{exp} group. Two singleton class--\texttt{exp} strata are excluded only from the second sensitivity. These analyses examine sensitivity to class, location, and repository-group structure. They were not used to define or revise the primary external decision rule.

\subsection{Exploratory defect covariates}
\label{sec:covariates}

Defect area is the positive-pixel count in the native mask. Local contrast is computed on the native 8-bit grayscale image as
\begin{equation}
 c_i=\left|\operatorname{mean}(I[M_i])-
 \operatorname{median}(I[R_i])\right|,
 \qquad R_i=\operatorname{dilate}_{5\ \mathrm{pixels}}(M_i)\setminus M_i.
 \label{eq:contrast}
\end{equation}
The models use standardized $\log(a_i)$ and $\log(c_i+1/255)$. The originally planned logistic specifications include detector, modality, seed, detector--modality terms, and area/contrast interactions. Sandwich standard errors are clustered by image path~\citep{cameron2015cluster}. After the crossed-mask analysis, we retained these models as diagnostics of the geometric component of raw overlap, not as estimates of a physical minimum detectable area or a defect-specific localization effect.

\section{Results}
\label{sec:results}

\subsection{Image-level performance at the calibrated threshold}

Across the nine primary ISP-AD cells, the mean FPR on independent normal images ranges from 4.8\% to 6.8\%, below the nominal 10\% target (Table~\ref{tab:main-results}). These values describe the evaluation splits and do not estimate FPR after a future distribution shift. DINOv2--\LSMtwo illustrates the difference between ranking and operating-point performance: its mean AUROC is 0.974 and its point-defect alarm recall is 70.7\%. Across seeds, AUROC varies by approximately 0.006, whereas recall ranges from 45.5\% to 87.9\%.

\begin{table}[H]
\centering
\caption{Primary ISP-AD results over three seeds. Alarm+overlap is the uncorrected diagonal event in Eq.~\eqref{eq:compatibility}. Overlap given alarm is descriptive because it has not been adjusted using the crossed baseline.}
\label{tab:main-results}
\resizebox{\textwidth}{!}{%
\begin{tabular}{llrrrrr}
\toprule
Detector & Modality & AUROC & FPR & Alarm recall & Alarm+overlap & Overlap $\mid$ alarm \\
\midrule
PatchCore & \LSMtwo & 0.893 & 6.7\% & 26.3\% (26/99) & 23.2\% (23/99) & 88.5\% \\
PatchCore & \LSMone & 0.752 & 6.5\% & 13.0\% (18/138) & 10.9\% (15/138) & 83.3\% \\
PatchCore & \ASM    & 0.860 & 6.8\% & 35.0\% (43/123) & 2.4\% (3/123) & 7.0\% \\
\addlinespace
DINOv2 & \LSMtwo & 0.974 & 6.8\% & 70.7\% (70/99) & 68.7\% (68/99) & 97.1\% \\
DINOv2 & \LSMone & 0.927 & 6.7\% & 56.5\% (78/138) & 32.6\% (45/138) & 57.7\% \\
DINOv2 & \ASM    & 0.942 & 5.3\% & 70.7\% (87/123) & 38.2\% (47/123) & 54.0\% \\
\addlinespace
PaDiM-style & \LSMtwo & 0.923 & 5.8\% & 31.3\% (31/99) & 16.2\% (16/99) & 51.6\% \\
PaDiM-style & \LSMone & 0.633 & 6.8\% & 10.1\% (14/138) & 9.4\% (13/138) & 92.9\% \\
PaDiM-style & \ASM    & 0.802 & 4.8\% & 8.9\% (11/123) & 0.0\% (0/123) & 0.0\% \\
\bottomrule
\end{tabular}}
\end{table}

\subsection{Paired and crossed spatial overlap on ISP-AD}

Of 378 alarms, 230 intersect the corresponding defect mask, giving a raw overlap-given-alarm rate of 60.8\%. After comparison with crossed masks, eight of the nine detector--modality cells have a 95\% bootstrap interval whose lower endpoint is zero or negative (Table~\ref{tab:isp-lift}). DINOv2--\LSMtwo provides a clear example: its paired rate is 68.7\%, its crossed rate is 68.0\%, and the resulting lift is 0.007 ($p=0.769$). Pairing its score maps with masks from other \LSMtwo images therefore reproduces almost all of the raw overlap rate.

DINOv2--\ASM is the only cell with a strictly positive bootstrap lower bound after the nine-cell comparison: the paired rate is 0.382, the crossed rate is 0.123, and the lift is 0.259 (95\% CI 0.159--0.347; $p=10^{-5}$). Detector-blocked permutation tests aggregate the evidence across modalities and give $p=0.00373$ for PatchCore, $p=10^{-5}$ for DINOv2, and $p=0.91031$ for PaDiM. The prespecified ISP-AD criterion, which requires positive lower bounds in at least seven cells and $p<0.01$ for all three detectors, is not met.

\begin{table}[H]
\centering
\caption{ISP-AD chance-corrected spatial evidence. Rates are unconditional alarm-and-overlap probabilities after averaging three seed matrices. Cellwise $p$ values are one-sided and interpreted against $0.05/9=0.0056$.}
\label{tab:isp-lift}
\footnotesize
\begin{tabular*}{\textwidth}{@{\extracolsep{\fill}}llrrrrr@{}}
\toprule
Detector & Modality & Paired & Crossed & Lift $\Delta$ & Bootstrap 95\% CI & Permutation $p$ \\
\midrule
PatchCore & \LSMtwo & 0.232 & 0.225 & 0.007 & [0.000, 0.025] & 0.75947 \\
PatchCore & \LSMone & 0.109 & 0.097 & 0.012 & [$-0.002$, 0.033] & 0.19920 \\
PatchCore & \ASM    & 0.024 & 0.000 & 0.024 & [0.000, 0.070] & 0.02416 \\
\addlinespace
DINOv2 & \LSMtwo & 0.687 & 0.680 & 0.007 & [0.000, 0.026] & 0.76862 \\
DINOv2 & \LSMone & 0.326 & 0.303 & 0.023 & [$-0.002$, 0.066] & 0.02011 \\
DINOv2 & \ASM    & 0.382 & 0.123 & 0.259 & [0.159, 0.347] & 0.00001 \\
\addlinespace
PaDiM-style & \LSMtwo & 0.162 & 0.160 & 0.002 & [0.000, 0.009] & 0.94069 \\
PaDiM-style & \LSMone & 0.094 & 0.094 & 0.001 & [0.000, 0.003] & 0.95699 \\
PaDiM-style & \ASM    & 0.000 & 0.006 & $-0.006$ & [$-0.015$, 0.000] & 1.00000 \\
\bottomrule
\end{tabular*}
\end{table}

\begin{figure}[H]
\centering
\includegraphics[width=0.98\textwidth]{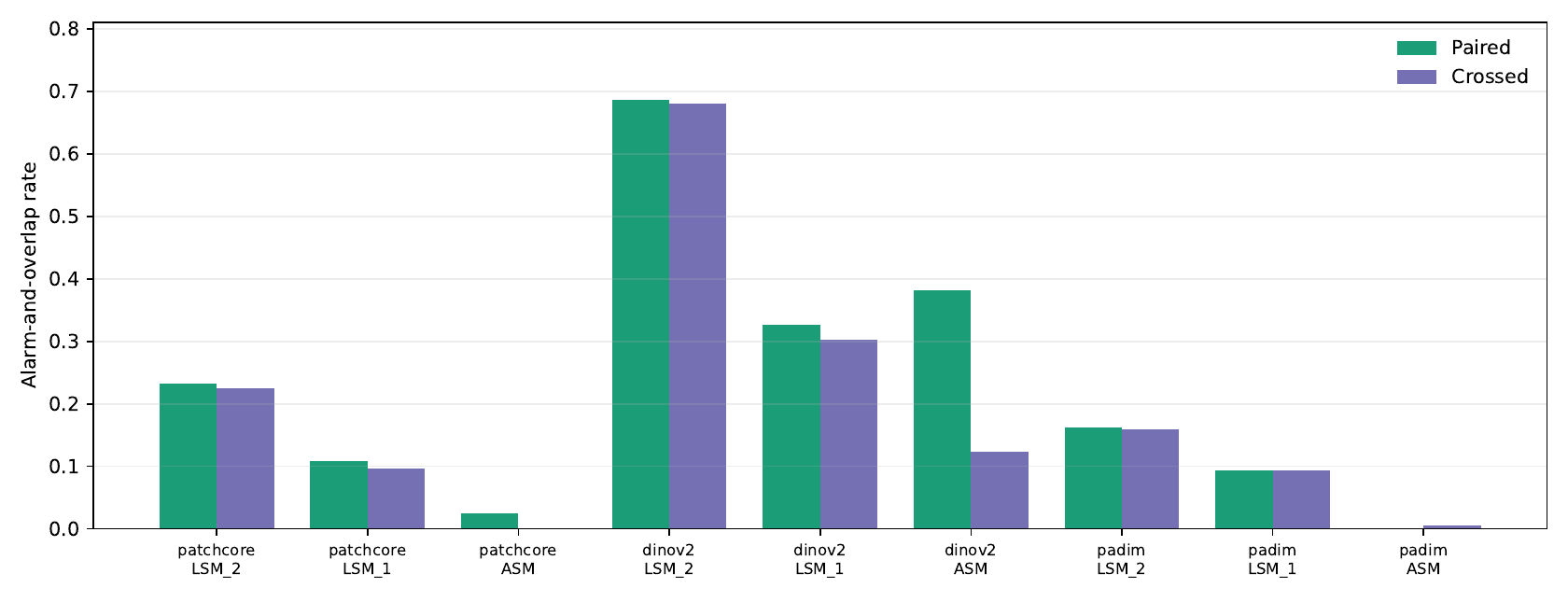}
\caption{Paired and crossed alarm-and-overlap rates on ISP-AD. Near-equal bars show cells in which raw mask overlap is largely reproduced after image--mask identity is broken. Comparisons are made within detector and modality; Table~\ref{tab:isp-lift} reports the numerical lifts and bootstrap intervals.}
\label{fig:isp-crossed}
\end{figure}

\subsection{External evaluation on Magnetic Tile Defect}

The paired and crossed rates are clearly separated on Magnetic Tile Defect for both representation families (Table~\ref{tab:mtd}). WRN50 patch memory has a paired rate of 0.260 and a crossed rate of 0.058, giving a lift of 0.203 (95\% CI 0.169--0.236). For ViT-B/16, the corresponding rates are 0.301 and 0.069, and the lift is 0.231 (0.202--0.262). Both one-sided permutation tests reach the Monte Carlo minimum of $p=10^{-5}$ and satisfy the prespecified two-detector correction.

At this operating point, the differences amount to approximately 20 additional alarm-and-overlap events per 100 evaluated defect images for WRN50 and 23 per 100 for ViT-B/16, relative to the empirical crossed-mask reference. This population-level interpretation does not classify individual overlaps as causal defect responses.

At image level, mean AUROC is 0.743 for WRN50 and 0.771 for ViT-B/16, while mean alarm recall is 37.2\% and 36.5\%, respectively. Mean FPR is 7.3\% for WRN50 and 7.5\% for ViT-B/16, with seed ranges of 6.1--9.2\% and 4.8--12.3\%. One ViT-B/16 seed exceeds the 10\% target; Appendix~\ref{app:reproducibility} reports the seed-specific values. The calibration guarantee applies over repeated calibration samples and permits individual splits to exceed the target, so threshold variability is reported separately from spatial evidence.

\begin{table}[H]
\centering
\caption{Primary Magnetic Tile evaluation over three seeds. Entries are means; FPR also shows the seed range. Image metrics include all 392 defects; spatial columns use 388 nonempty masks.}
\label{tab:mtd}
\resizebox{\textwidth}{!}{%
\begin{tabular}{lrrrrrrrr}
\toprule
Detector & AUROC & FPR & Alarm recall & Paired & Crossed & Lift & Bootstrap 95\% CI & $p$ \\
\midrule
WRN50 patch memory & 0.743 & 7.3\% [6.1, 9.2] & 37.2\% & 0.260 & 0.058 & 0.203 & [0.169, 0.236] & 0.00001 \\
ViT-B/16 patch memory & 0.771 & 7.5\% [4.8, 12.3] & 36.5\% & 0.301 & 0.069 & 0.231 & [0.202, 0.262] & 0.00001 \\
\bottomrule
\end{tabular}}
\end{table}

\begin{figure}[H]
\centering
\includegraphics[width=0.76\textwidth]{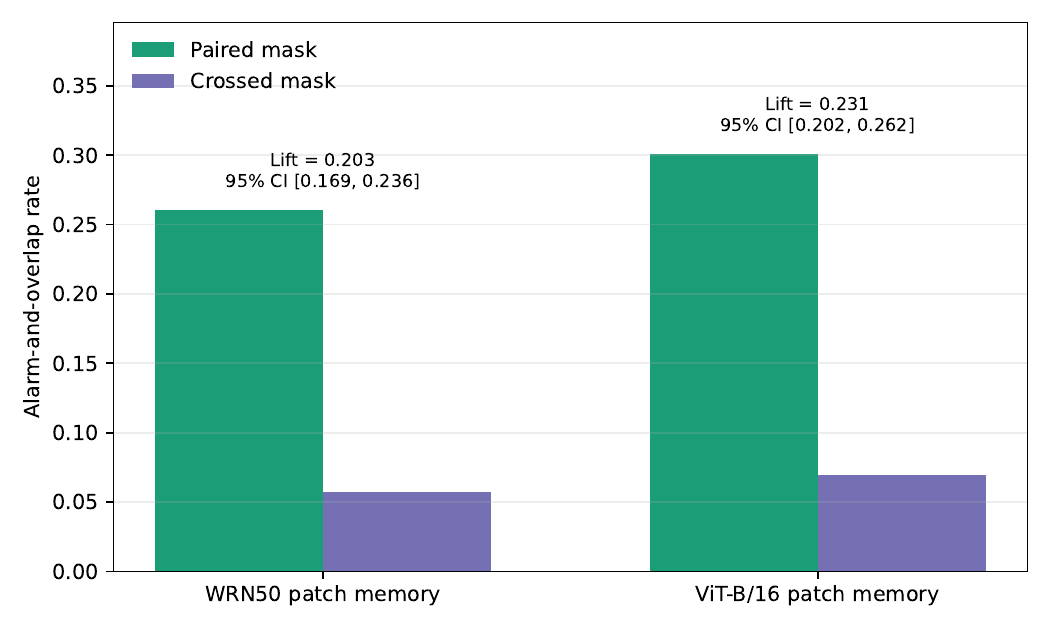}
\caption{Magnetic Tile paired and crossed rates under the protocol specified before result inspection. Annotations give paired-minus-crossed lift with its percentile bootstrap 95\% confidence interval.}
\label{fig:mtd-crossed}
\end{figure}

The post-hoc sensitivity analyses give the same qualitative result (Table~\ref{tab:mtd-stratified}). Restricting crossed masks to the same defect class increases the crossed rates and leaves lifts of 0.185 for WRN50 and 0.210 for ViT-B/16. Restricting them by both class and the filename-defined \texttt{exp} group leaves lifts of 0.199 and 0.225 among 386 eligible defects. All intervals remain above zero, and all within-stratum permutation tests reach the Monte Carlo minimum. Because the restrictions were selected after the pooled analysis, these results are robustness checks rather than independent confirmation.

\begin{table}[H]
\centering
\caption{Post-hoc Magnetic Tile stratified-crossing sensitivity. Class--\texttt{exp} analysis excludes two singleton strata. Pooled rows reproduce the primary analysis; stratified rows use within-stratum bootstrap and permutation.}
\label{tab:mtd-stratified}
\footnotesize
\begin{tabular}{llrrrrr}
\toprule
Detector & Crossing set & Spatial $n$ & Paired & Crossed & Lift (95\% CI) & $p$ \\
\midrule
WRN50 & pooled (primary) & 388 & 0.260 & 0.058 & 0.203 [0.169, 0.236] & 0.00001 \\
WRN50 & within class & 388 & 0.260 & 0.075 & 0.185 [0.154, 0.211] & 0.00001 \\
WRN50 & within class--\texttt{exp} & 386 & 0.262 & 0.063 & 0.199 [0.153, 0.208] & 0.00001 \\
\addlinespace
ViT-B/16 & pooled (primary) & 388 & 0.301 & 0.069 & 0.231 [0.202, 0.262] & 0.00001 \\
ViT-B/16 & within class & 388 & 0.301 & 0.091 & 0.210 [0.182, 0.230] & 0.00001 \\
ViT-B/16 & within class--\texttt{exp} & 386 & 0.301 & 0.076 & 0.225 [0.180, 0.232] & 0.00001 \\
\bottomrule
\end{tabular}
\end{table}

\subsection{Sensitivity of raw-overlap models to mask geometry}

The image-alarm model gives a standardized log-area coefficient of 1.94 (cluster-robust 95\% CI 1.01--2.87). Replacing the response with the raw alarm-and-overlap event increases the coefficient to 3.69 (2.37--5.01). Geometry contributes to this increase because larger masks and their one-cell dilations offer more opportunities to intersect a selected contributor, including under crossed image--mask pairing. We therefore use the coefficient difference as a geometry diagnostic and do not interpret raw-overlap $a_{50}$ or $a_{90}$ values as physical detection limits.

The standardized log-contrast coefficient is 2.54 (0.18--4.90) for the alarm response and 3.56 ($-0.55$ to 7.68) for raw alarm-and-overlap. The latter interval crosses zero. Equation~\eqref{eq:contrast} specifies the five-pixel reference ring and the use of a background median. At ring widths of 3, 5, and 9 pixels, the alarm-model coefficients are 2.81 (0.36--5.27), 2.54 (0.18--4.90), and 2.04 ($-0.10$ to 4.18). The corresponding raw-overlap coefficients are 3.54 ($-0.57$ to 7.66), 3.56 ($-0.55$ to 7.68), and 2.96 ($-0.73$ to 6.65). The point estimate decreases as the reference ring widens, and the raw-overlap interval includes zero at all three widths.

\subsection{Calibration sample size and deployment implications}

Table~\ref{tab:fpr-planning} gives exact finite-sample limits. With $m=150$ and the rank reselected for each target, 10\%, 5\%, and 2\% FPR claims are supported; the 2\% setting requires the sample maximum. The smallest target supported by any finite order statistic is
\begin{equation}
  \alpha_{\min}=1-\delta^{1/m}=0.01977.
\end{equation}
If the original rank $r=142$ is held fixed rather than reselected, the smallest 95\%-confidence target is 9.417\%. A 1\% target is unsupported: even the sample maximum has only 77.85\% calibration confidence. The minimum normal count for any finite threshold is
\begin{equation}
  m_{\min}=\left\lceil\frac{\log\delta}{\log(1-\alpha)}\right\rceil.
\end{equation}

\begin{table}[H]
\centering
\caption{One-sided distribution-free sample planning at $1-\delta=0.95$. ``Unsupported'' means that the sample maximum cannot reach 95\% confidence with 150 normals.}
\label{tab:fpr-planning}
\begin{tabular}{rrrrr}
\toprule
Target FPR $\alpha$ & Rank for $m=150$ & Achieved confidence & Minimum $m$ & Status at $m=150$ \\
\midrule
10\%  & 142 & 96.93\% & 29   & supported \\
5\%   & 148 & 98.18\% & 59   & supported \\
2\%   & 150 & 95.17\% & 149  & supported \\
1\%   & --  & 77.85\% maximum & 299  & unsupported \\
0.5\% & --  & 52.85\% maximum & 598  & unsupported \\
0.1\% & --  & 13.94\% maximum & 2995 & unsupported \\
\bottomrule
\end{tabular}
\end{table}

The 10\% setting is a benchmark operating point, not a general production specification. Its practical acceptability depends on throughput, the inspection stage, and the cost of false alarms. A 95\%-confidence claim near 1\% FPR requires at least 299 representative and effectively independent normal calibration units, even before threshold stability, drift, and economic costs are considered.

These experiments do not determine a universal recalibration interval. In deployment, recalibration should be considered after changes to the detector, camera, illumination, tooling, material, or production process, or when monitoring indicates a shift in the normal-score distribution. The new threshold should be based on a calibration sample that represents the resulting operating condition.

\section{Discussion}
\label{sec:discussion}

\subsection{Interpreting the reported metrics}

The reported metrics answer related but distinct questions (Table~\ref{tab:claim-hierarchy}). AUROC measures whether anomaly scores tend to rank above normal scores across thresholds. Alarm recall measures how often defects exceed a selected threshold whose normal-tail behavior has been calibrated. Raw alarm-and-overlap records intersection with the annotated mask, whereas spatial-evidence lift measures how much that intersection exceeds the crossed-mask reference rate. Each calculation requires finite scores from the specified pipeline.

\begin{table}[H]
\centering
\caption{Interpretation of the reported evidence. Each metric requires information that is not supplied by the preceding metric alone.}
\label{tab:claim-hierarchy}
\footnotesize
\begin{tabular}{@{}>{\raggedright\arraybackslash}p{0.20\textwidth}>{\raggedright\arraybackslash}p{0.37\textwidth}>{\raggedright\arraybackslash}p{0.34\textwidth}@{}}
\toprule
Evaluation level & Required evidence & Not determined by \\
\midrule
Score validity & finite required scores; valid/attempted runs & summaries over valid runs only \\
Ranking & image AUROC on held-out normals and defects & a valid score alone \\
Operating performance & fixed threshold, FPR statement, alarm recall & AUROC \\
Raw compatibility & alarm and true-mask intersection & image alarm \\
Spatial specificity & positive paired-minus-crossed lift with uncertainty & raw overlap \\
\bottomrule
\end{tabular}
\end{table}

The DINOv2--\LSMtwo result shows why these quantities should remain separate. Its raw overlap-given-alarm rate is 97.1\%, but the unconditional paired and crossed rates are almost identical. On Magnetic Tile Defect, the paired rate remains well above the crossed rate for both feature representations. The crossed comparison tests whether overlap is specific to the matched image and mask; it does not replace conventional image- or region-level localization metrics.

\subsection{Scope of the crossed baseline}

The crossed reference retains the empirical number and location of score contributors, alarm frequency, mask size, mask dilation, and recurrent mask position. For images with strong spatial structure, it provides a closer reference than uniformly sampled pixels. Class- and filename-group matching can further restrict the comparison when suitable metadata are available. Even then, no negative-sampling rule removes every spatial bias~\citep{jia2020fnaauc}. The test does not identify a causal visual mechanism, assess mask-boundary accuracy, or measure segmentation completeness. It also has little power to distinguish defects that share both appearance and position. Region-level precision and recall may be more informative for large defects; the crossed test is primarily a specificity check for small, localized defects.

The area analysis provides an example of the geometric effect. An intersection outcome includes mask area by construction, so its area coefficient can increase even when physical sensitivity has not changed. Future studies could estimate lift for individual masks or construct location-matched negative masks. Such analyses would be most informative if their matching rules were specified before the results were examined.

\subsection{Practical interpretation of the tolerance threshold}

The order-statistic threshold is distribution free with respect to the shape of the normal-score distribution, but its validity still depends on the sampling design. Calibration scores must represent effectively independent draws from the future normal population, and the score pipeline must remain fixed. Images correlated by batch, tool state, or repeated specimen should be calibrated and analyzed at the corresponding independent unit. A distribution shift requires monitoring and, when necessary, recalibration; MVTec AD 2 illustrates how static thresholds can fail under changed imaging conditions~\citep{hecklerkram2026mvtecad2}.

Sample planning is therefore part of the operating specification. With 150 normal images, lowering the nominal target from 10\% to 1\% cannot produce a 95\%-confidence guarantee at 1\% FPR. Additional data are needed to support the lower target and to study drift, threshold stability, and economic cost. An evaluation split may nevertheless exceed the target because the guarantee is defined over repeated calibration samples and allows failure probability $\delta$.

\subsection{Limitations}
\par\smallskip

The empirical study uses two public industrial datasets rather than multi-site production streams. The crossed comparison was developed on ISP-AD, after which the Magnetic Tile protocol was fixed before its results were inspected. Magnetic Tile thus provides an independent-dataset evaluation of the revised procedure, but not a preregistration of the complete study. ISP-AD contains 120 modality-specific point-defect images, with no verified key for linking physical specimens across modalities. Magnetic Tile filenames contain \texttt{exp} groups, but the repository does not state whether these groups correspond to independent production units. The FPR statement therefore depends on image-level independent sampling and should not be interpreted as batch-robust.

Several measurement choices are specific to this study. The one-cell dilation and contributor reducers depend on the detector grid. The crossed baseline adjusts for chance intersection under these choices, but other dilation or contributor rules may give different results. The covariate model is high dimensional relative to the 120 image clusters, and its raw-overlap response remains sensitive to mask geometry; its coefficients are therefore not interpreted as physical capability limits. The EfficientAD check in Appendix~\ref{app:efficientad} uses a 10,000-step compute-adapted run rather than the official 70,000-step schedule. The analysis also excludes economic costs, severity labels, temporal drift, and dependence among sequential production images. The 10\% FPR is suitable as a benchmark operating point or a human-reviewed screening stage, but not as a general production recommendation.

\section{Conclusion}
\label{sec:conclusion}

This study combines a distribution-free tolerance threshold with a paired-minus-crossed spatial test for industrial anomaly detection. The threshold provides a finite-sample statement about the normal-tail probability at a selected operating point, while the spatial test estimates how much matched image--mask overlap exceeds the dataset's crossed-mask rate. On ISP-AD, eight of nine detector--modality cells have no positive bootstrap lower bound for spatial-evidence lift despite 230 raw overlaps. On Magnetic Tile Defect, both feature representations show positive lift, which remains positive in class- and filename-group-matched sensitivity analyses. Exact sample planning also shows that 150 calibration normals cannot support a 95\%-confidence claim at 1\% FPR. Reporting these quantities separately gives a clearer account of ranking performance, threshold behavior, and the spatial specificity of detector responses.

\appendix

\section{Tolerance-limit proof with ties}
\label{app:proof}

Let $F$ be the target-normal score distribution and $q=\inf\{x:F(x)\geq1-\alpha\}$. Then $P_0(S>q)\leq\alpha$ and $P_0(S\geq q)\geq\alpha$. If at least $k=m-r+1$ calibration scores are at least $q$, then $S_{(r)}\geq q$ and consequently $P_0(S>S_{(r)})\leq\alpha$. Under iid calibration, the count of scores at least $q$ is binomial with success probability $p\geq\alpha$. Its upper-tail probability is therefore at least the $\mathrm{Binomial}(m,\alpha)$ tail in Eq.~\eqref{eq:binomial}. This proves Eq.~\eqref{eq:guarantee}. Atoms and ties make the strict decision conservative; a continuity assumption is unnecessary.

\section{Reproducibility checklist}
\label{app:reproducibility}

\begin{table}[H]
\centering
\caption{Seed-specific Magnetic Tile false-positive estimates on 786 independent normal images per split. Intervals are two-sided 95\% Wilson intervals and quantify evaluation-set sampling uncertainty, not the separate calibration-sample coverage in Eq.~\eqref{eq:guarantee}.}
\label{tab:mtd-fpr-seeds}
\begin{tabular}{llr@{\hspace{2em}}rr}
\toprule
Detector & Seed & False positives & FPR & Wilson 95\% CI \\
\midrule
WRN50 & 20260801 & 52 & 6.62\% & [5.08\%, 8.57\%] \\
WRN50 & 20260802 & 48 & 6.11\% & [4.64\%, 8.00\%] \\
WRN50 & 20260803 & 72 & 9.16\% & [7.34\%, 11.38\%] \\
\addlinespace
ViT-B/16 & 20260801 & 97 & 12.34\% & [10.22\%, 14.83\%] \\
ViT-B/16 & 20260802 & 38 & 4.83\% & [3.54\%, 6.57\%] \\
ViT-B/16 & 20260803 & 41 & 5.22\% & [3.87\%, 7.00\%] \\
\bottomrule
\end{tabular}
\end{table}

\begin{enumerate}
  \item ISP-AD archive: public unsupervised split with original filenames; Magnetic Tile: original JPG/PNG pairs.
  \item Seeds: 20260801, 20260802, and 20260803.
  \item Primary threshold: $\alpha=0.10$, $\delta=0.05$, $m=150$, $r=142$, strict $S>\tau$.
  \item Primary ISP contributors: PatchCore maximum cell; DINOv2 and PaDiM top 1\%; masks pooled to the native grid and dilated by one cell.
  \item Crossed baseline: all off-diagonal mask--map pairings within detector and modality; seed matrices averaged before image-level inference. Post-hoc external sensitivities restrict crossing to defect class and class--\texttt{exp} strata.
  \item Uncertainty: 5,000 paired-image bootstrap replicates; 99,999 one-sided assignment permutations.
  \item Local-contrast sensitivity: native-image rings of 3, 5, and 9 pixels; the five-pixel ring is primary.
  \item External protocol: 16 support normals, 150 calibration normals, all remaining normals, all defects, WRN50 and ViT-B/16 maximum-distance patch memories; four empty masks excluded only from spatial inference.
  \item Inference unit: full defect-image path. No cross-modality merging is assumed.
  \item EfficientAD tiers: \LSMtwo development; \LSMone and \ASM fixed-protocol held-modality. The failed \ASM seed remains in the attempted denominator.
\end{enumerate}

\section{Secondary EfficientAD score-validity check}
\label{app:efficientad}

After completing the three-detector ISP-AD analysis, we evaluated EfficientAD-S from Anomalib 2.4.2 as a secondary check of score validity. The configuration used the official small-teacher weights, ImageNette penalty images, batch size one, $256\times256$ inputs, an Adam learning rate of $10^{-4}$, weight decay of $10^{-5}$, and 10,000 training steps. The training budget was selected after an \LSMtwo development run. The \LSMone and \ASM results are treated as held-modality evaluations under the fixed protocol, not as prospective or external validation. Teacher statistics and branch-map percentiles were estimated from support normals, and threshold calibration was performed separately. Raw mask overlap from this check is not included in the chance-corrected spatial analysis.

Table~\ref{tab:efficientad} reports means over runs with defined scores and retains the valid/attempted denominator. The \ASM run for seed 20260802 completed training but did not produce a valid normalized anomaly score. The 90th and 99.5th percentiles of the support-normal student--teacher map were both 0.4036534727, making the denominator of the official affine normalization equal to zero. The raw branch maps remained finite, whereas the required normalized scores for calibration and evaluation were NaN or infinite. We report the run as undefined and do not replace the specified score with a post-hoc raw-map alternative.

\begin{table}[H]
\centering
\caption{Secondary EfficientAD-S-10k score-validity check. Means use runs with defined scores; the valid/attempted denominator includes the undefined run.}
\label{tab:efficientad}
\begin{tabular}{llrrrr}
\toprule
Modality & Evidence tier & Defined / attempted & AUROC & FPR & Point alarm recall \\
\midrule
\LSMtwo & development & 1/1 & 0.978 & 8.5\% & 97.0\% \\
\LSMone & fixed held-modality & 3/3 & 0.838 & 5.0\% & 41.3\% \\
\ASM    & fixed held-modality & 2/3 & 0.881 & 6.3\% & 52.4\% \\
\bottomrule
\end{tabular}
\end{table}

\begin{figure}[H]
\centering
\includegraphics[width=0.96\textwidth]{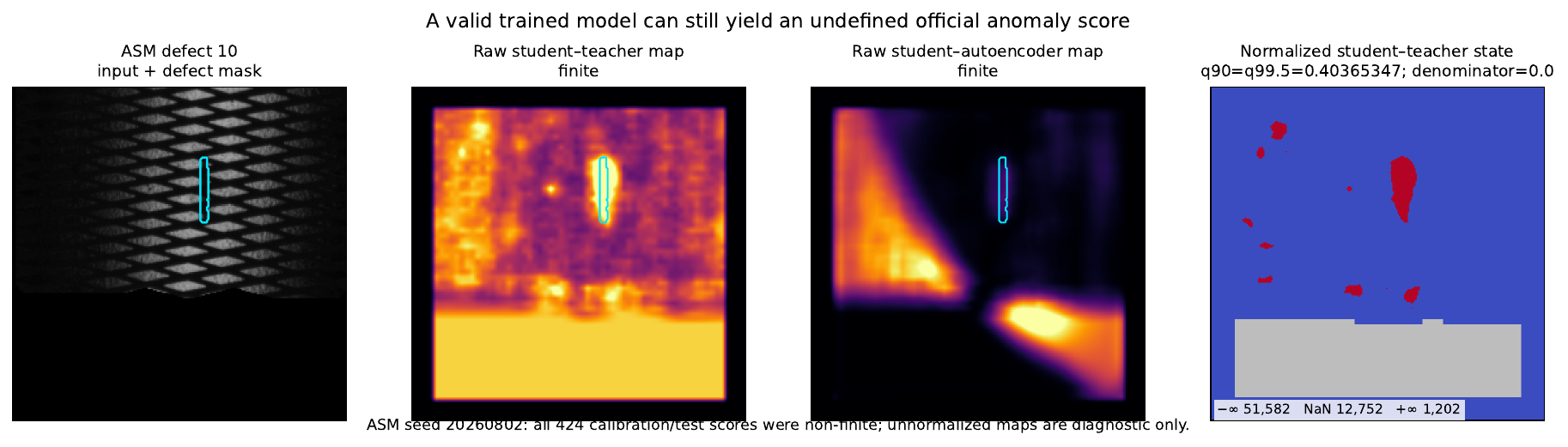}
\caption{The EfficientAD run with an undefined normalized score. The raw student--teacher and student--autoencoder maps are finite, but the affine normalization has zero range and all required normalized calibration and evaluation scores are undefined.}
\label{fig:failure}
\end{figure}

\section{Numerical failure-handling rule}
\label{app:failure-rule}

A run is valid only if every calibration and evaluation score required by the specified pipeline is finite. An undefined normalizer, non-finite score, missing checkpoint, or incomplete evaluation is reported with its dataset role and seed. Failed runs are not assigned zero AUROC, because AUROC is undefined, and remain in the valid/attempted denominator. Diagnostic alternative scores may be reported post hoc but do not replace the primary result.

\bibliographystyle{plainnat}
\bibliography{references}

\end{document}